\documentclass{article}

\usepackage[preprint]{corl_2026}

\usepackage{amsmath}
\usepackage{amssymb}
\usepackage{graphicx}
\usepackage{caption}
\usepackage{booktabs}
\usepackage{multirow}
\usepackage{hyperref}
\usepackage{cleveref}

\newcommand{\abbr}{SPD}

\newcommand{\algname}{\emph{Simulation Pre-training for Dexterity}}

\newcommand{\aff}[1]{\textsuperscript{\normalfont#1}}

\title{Pre-training Visual Dexterity in Simulation}
\author{%
  \begin{minipage}[t]{\dimexpr\linewidth-2\tabcolsep\relax}
    \centering\normalfont
    Sarthak Kamat\aff{1}\thanks{Equal contribution. Correspondence to:
      \texttt{sartk@cs.stanford.edu}, \texttt{abrashid@mit.edu}.\\
      \aff{1}Stanford University \quad
      \aff{2}MIT \quad
      \aff{3}Scale AI}\quad
    Adam Rashid\aff{2}\footnotemark[1]\quad 
    Satvik Sharma\aff{1}\quad 
    Aseem Doriwala\aff{3}\quad \\
    Chelsea Finn\aff{1}\quad
    Phillip Isola\aff{2}\quad
    C.~Karen Liu\aff{1}\\[6pt]
    \href{https://spd.bot}{\texttt{spd.bot}}%
  \end{minipage}%
}
\begin{document}
\maketitle
\vspace{-2em}

\begin{abstract}
Large-scale pre-training has made robot policy fine-tuning increasingly data-efficient, but this progress has largely been driven by datasets and embodiments built around simple parallel-jaw grippers. Dexterous, multi-fingered hands remain comparatively data-starved because real teleoperation is costly to scale, while human hand video is off-embodiment and requires lossy pose estimation and retargeting. We introduce \algname~(\abbr), a pre-training framework for dexterous manipulation that uses data entirely collected in simulation. In \abbr, humans manipulate virtual objects inside a VR headset, enabling on-embodiment trajectories and robot-free collection. With the help of five operators, we collect 75 hours of multi-task dexterous manipulation over one week, and use it to pre-train a causal transformer on a sequence modeling objective. We study the benefits of simulation pre-training on real-world tasks by fine-tuning on 1--2 hours of physical demonstrations on a 56-DoF bimanual dexterous setup. We find that our approach outperforms training behavior cloning policies from scratch, showing that simulation teleoperation is a viable pre-training source for real-world dexterous manipulation. We perform ablation studies, measuring the benefits of history conditioning and short action chunks for reactive control.
\end{abstract}

\begin{figure}[t]
    \centering
    \includegraphics[width=1\linewidth]{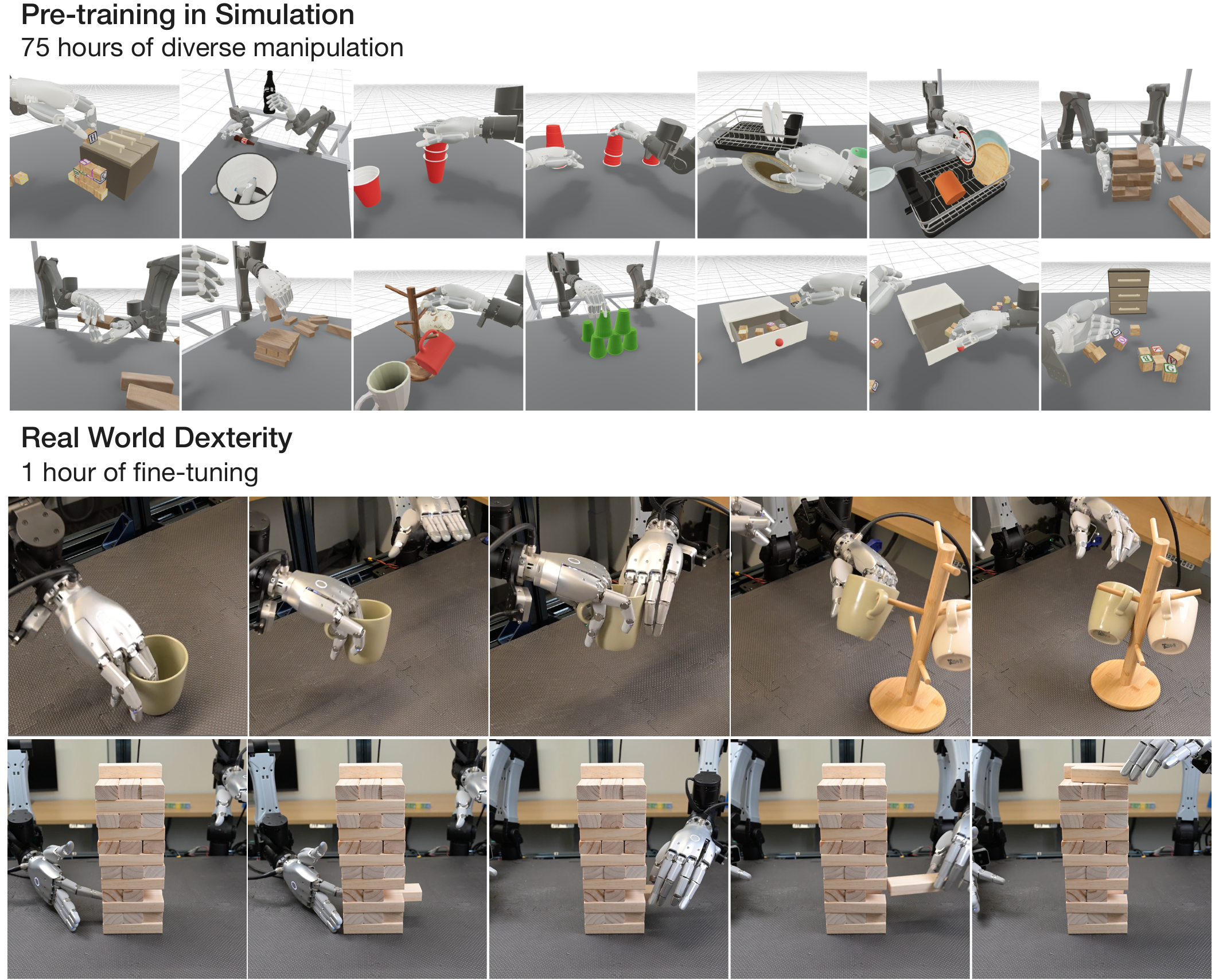}
    \caption{\textbf{Simulation pre-training for dexterity.} Top: 75 hours of teleoperated simulation data spanning six scenes and hundreds of object instances, covering contact-rich low-level dexterous behavior. Bottom: the pre-trained policy adapts to the real-world dexterous tasks shown, with 1 hour of real demonstrations.
    }
    
    \label{fig:teaser}
\end{figure}

\section{Introduction}
\label{sec:introduction}

Dexterous manipulation is a challenging problem in robotics because it demands precise control during periods of high contact and partial observability, with low tolerance for latency. Behavior cloning has been shown to be an effective  technique for training bimanual dexterous policies \citep{aloha-unleashed}, but performance is limited by the quantity and coverage of task demonstrations. To address this, recent work pre-trains policies on large multi-task datasets, before fine-tuning on a smaller set of demonstrations of the target task~\citep{pi0, lbmtri2025, gen1, abc2026}. This amortizes the cost of learning broadly useful visuomotor representations and makes policy learning data-efficient.

Teleoperation data is the ideal source for pre-training as it has no train-test divergence, but is challenging for multi-fingered hands because their cost and fragility limit data throughput. Handheld interfaces, such as UMI, enable scalable data collection without a robot \citep{umi, gen1, sunday2026act2preview}, but are also difficult to
extend to hands, as the device must avoid an embodiment gap while giving the operator control over every degree of freedom. Human hand video is abundant and cheap, but extracting robot-usable supervision from it is difficult because contact-rich hand-object interaction involves occlusion, resulting in noisy hand pose reconstruction. Even when human hands can be recorded accurately, such as by wearing motion-capture gloves, there remain differences in contact points and finger actuation, resulting in demonstrations that cannot be faithfully executed on the robot.

In this work, we study whether 
data collected in physics simulators can enable scalable pre-training for dexterous hands.  Our framework, \algname{} (\abbr{}), consists of two teleoperation systems: \texttt{spd-vr}, virtual reality  software used to collect pre-training data, and \texttt{spd-teleop}, a real-world teleoperation system used to collect  post-training data. The two systems are aligned, sharing top and
wrist camera placements, an identical pair of dexterous hands and 6-DoF arms, and a
similar collection of objects. Since our VR data collection has no dependence on physical hardware, it enables fast resets of virtual objects, parallel operation, and decentralized collection.

Using \texttt{spd-vr}, five operators collect 75 hours of long-horizon,
multi-task demonstrations over one week, spanning six scenes and hundreds
of object instances. We pre-train a diffusion transformer policy
on this dataset and fine-tune it on each
downstream task with 1--2 hours of robot data. Across five
real-world tasks --- plate racking, mug hanging, playing Jenga, cup
stacking, and tossing bottles into a bin --- pre-training on simulation data outperforms training from scratch.
Our ablations further show the value of conditioning on visuomotor
history. With context, the policy can re-plan its actions frequently,
staying reactive without losing temporal coherence, and this configuration
benefits the most from pre-training.

Beyond our experimental findings, we release our  pre-training dataset (\texttt{spd-75h}), VR teleoperation software (\texttt{spd-vr}), and six curated scenes with tuned contact parameters to support future
work on simulation pre-training for dexterity.

\section{Related Work}
\label{sec:related_work}

\textbf{Pre-training for robot policies.}
Multi-task imitation learning has become a dominant paradigm for robot policy pre-training, with large robot datasets \citep{robonet, bridgedatav2, openx, droid} and ALOHA-style teleoperation datasets \citep{aloha} enabling policies to learn from broad task and scene diversity. These datasets have supported increasingly general manipulation policies, including multi-task imitation and transformer-based policies \citep{mtopt, bcz, peract, rt1, octo, rdt, rpt}. Recent vision-language-action models extend this paradigm by adapting pre-trained vision-language backbones for action prediction \citep{rt2, openvla, pi0, molmoact, robocat}. Beyond zero-shot generalization, these works have shown that large-scale pre-trained policies can serve as useful priors for downstream fine-tuning across new tasks, embodiments, modalities, and action spaces \citep{octo, openvla, openvlaoft, rdt, robocat}. However, most of this progress has focused on manipulation with parallel-jaw grippers, whereas \textit{SPD} studies whether simulation pre-training can be extended to contact-rich manipulation with dexterous hands.

\textbf{Learning from humans.}
Because real-world robot teleoperation is expensive to scale, prior work has explored off-robot data sources that collect useful manipulation supervision without continuously operating a physical robot. One line of work uses handheld or portable interfaces to preserve action labels while reducing robot dependence \citep{dexumi, tacumi, umi3d, dexmouse, freetacman}. Another line of work pre-trains from human video or internet-scale visual data, either through learned representations \citep{r3m, mvp, realmvp, voltron, liv, roboclip} or through more policy-oriented video pre-training methods \citep{gr1, lapa, egoscale, vitra, mpi}. These approaches are attractive because they turn abundant human activity into scalable supervision, but human videos generally lack robot action labels and require action inference, pose reconstruction, retargeting, or latent-action discovery before they can supervise robot policies. Handheld and wearable systems address some of these issues by producing action-labeled data, but they often require specialized hardware, exoskeletons, or visual post-processing, and bulky interfaces can limit the dexterity of the human demonstrator. In contrast, \textit{SPD} uses simulation teleoperation to collect action-labeled data directly on the target dexterous-hand embodiment, avoiding post-hoc human-to-robot retargeting while retaining the scalability benefits of off-robot data collection.

\textbf{Simulation for dexterous manipulation.}
Simulation has been widely used for dexterous manipulation because it enables large-scale training for high-dimensional, contact-rich hands \citep{dactyl, rubikadr, dapg, dexmv, rpl}.
Subsequent sim-to-real systems have demonstrated increasingly capable dexterous skills using reinforcement learning with domain randomization from state-based and visual inputs \citep{dextreme, rotateit, dextrargb}.
Dexterous simulation benchmarks and training suites have expanded the scope of dexterous learning beyond individual hand-designed tasks, providing diverse objects, articulated environments, and multi-hand embodiments for training and evaluation \citep{dexart, bidexhands, dexpbt}.
In parallel-jaw and simple end-effector settings, recent sim-and-real co-training work shows that mixing real robot demonstrations with synthetic simulation data can improve vision-based manipulation policies, especially when the simulated data preserves task structure through digital cousins or controlled simulated demonstrations \citep{simrealcotrain}.
For dexterous robotic hands, simulation has often been used either with VR teleoperation systems for demonstration collection \citep{dart,lucidxr,iris}, or for RL-based sim-to-real transfer of task-specific skills \citep{dactyl,rubikadr,dextreme,visualdexterity_prior,dextrargb}. In contrast, \abbr{} uses simulation teleoperation as a broad pre-training data source: we collect diverse, action-labeled demonstrations directly on the target dexterous-hand embodiment and train a visuomotor policy intended to be fine-tuned across downstream real-world tasks, rather than optimized for a single simulated skill.

\section{Method}
\label{sec:method}
We study contact-rich visuomotor control with a pair of dexterous, multi-fingered robot hands. Our goal is to pre-train a policy on large-scale simulation data so that it can be adapted to downstream real-world dexterous manipulation tasks using only a small number of task-specific demonstrations. We describe our methodology for VR teleoperation (\Cref{sec:sim_data}), real-world teleoperation (\Cref{sec:real_data}), and policy learning (\Cref{sec:policy_learning}).

\subsection{VR Data Collection}
\label{sec:sim_data}

\begin{figure}[t]
    \centering
    \includegraphics[width=0.95\linewidth]{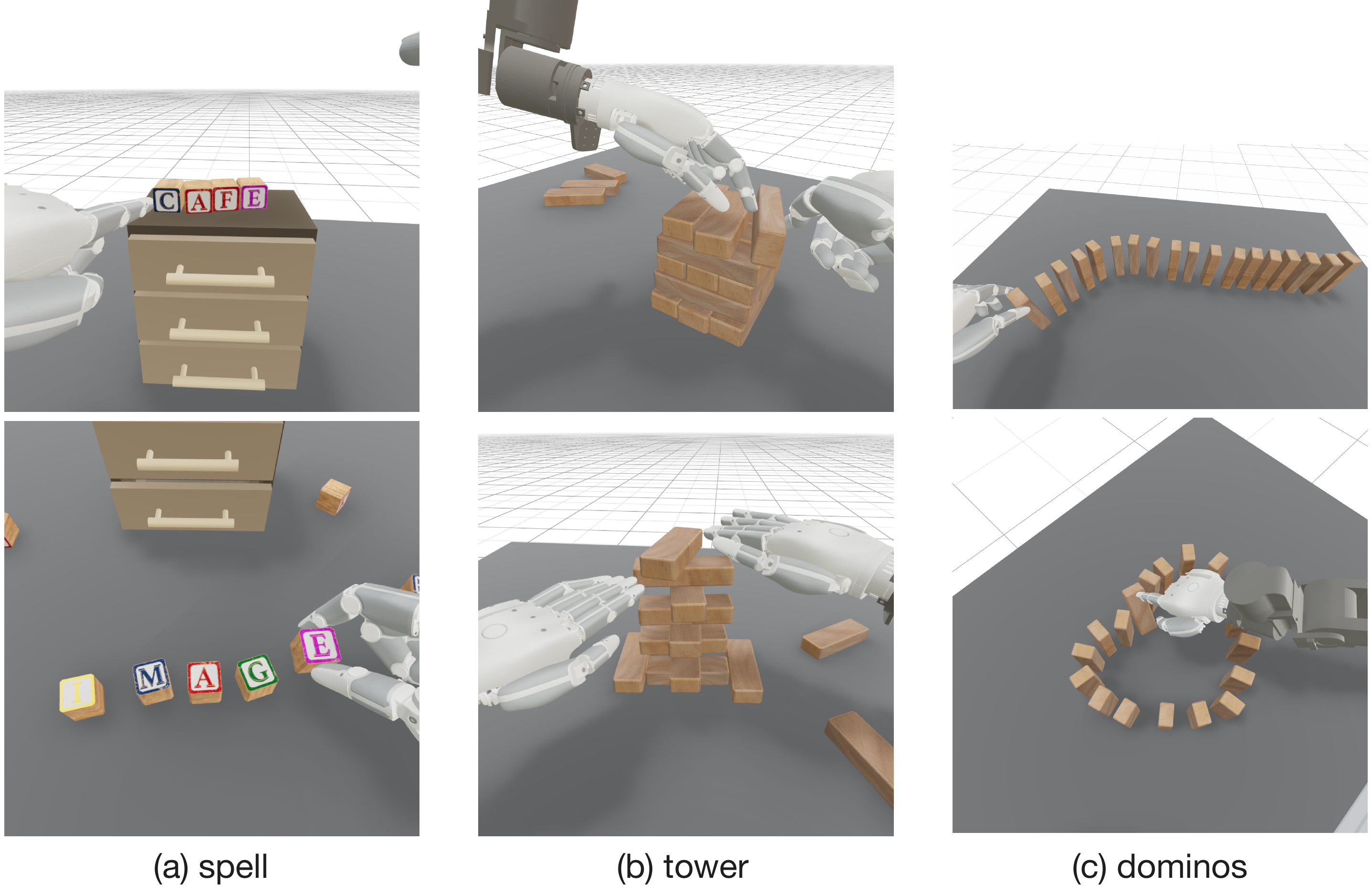}
    \caption{\textbf{Diverse operator strategies on open-ended simulation tasks.}
    Each pair of frames shows two distinct strategies the same operator pool produced on a single long-horizon task: (a) spelling a target word from a pile of letter blocks, (b) stacking blocks into a Jenga tower, and (c) arranging dominoes into a chain.
    }
    \label{fig:behaviors}
\end{figure}

During VR data collection, the operator controls the target bimanual dexterous robot directly in the MuJoCo physics simulator~\citep{todorov2012mujoco}. The simulation steps on a computer at 480 Hz, capturing hand poses from the cameras on the connected headset at 60 Hz. The detected wrist pose and fingertip positions are used to drive the simulated robot arms and hands through inverse kinematics. All objects are virtual, and hand-object contacts are physically simulated. The robot arms are made translucent to minimize occlusion during collection.

We collect data across six scenes: spelling blocks, dishes, mugs, bottles, cups, and Jenga bricks. On episode reset, we randomize a new task prompt, along with asset selection, physical properties, and initial positions of all objects.  
With the help of five operators, we collect approximately 2,000 episodes (75 hours) over one week. Tasks are long-horizon and open-ended, specifying the desired outcome without constraining strategy or subtask ordering, which yields diverse, contact-rich behavior across object instances, poses, and strategies (Figure~\ref{fig:behaviors}).

After data collection, we filter out extended periods of non-contact, and render our trajectories in parallel using an adapted version of the Madrona renderer~\citep{madrona}.  In order to increase visual randomization, we compute visual augmentations as GPU transforms applied after data-loading. During rendering, we save segmentation masks alongside the wrist-camera frames and use them to randomly tint object colors and swap background and table textures. We also perform a symmetry augmentation: swapping the two arms and reflecting the corresponding images, proprioception, and actions.

\subsection{Real-World Data Collection}
\label{sec:real_data}

Our hardware consists of two upgraded YAM Pro arms, each equipped with a 22-DoF Sharpa Wave dexterous hand. To reduce overheating due to the weight of the hands, we replace the 10:1 gear-ratio J3 and J4 motors on the YAM arms with 40:1. The setup includes three RealSense D405 cameras, a top camera mounted between the arms, and wrist cameras mounted on the ulnar side of each hand.

Real-world teleoperation mirrors our VR teleoperation for re-targeting and control, except we replace the headset's built-in hand tracking with Manus gloves for finger tracking and an attached Quest controller for wrist tracking. This produces smoother, more precise control, which we find necessary to control the robot through occlusions and at a distance.

\subsection{Policy Architecture and Training}
\label{sec:policy_learning}

\begin{figure}[t]
    \centering
    \includegraphics[width=\linewidth]{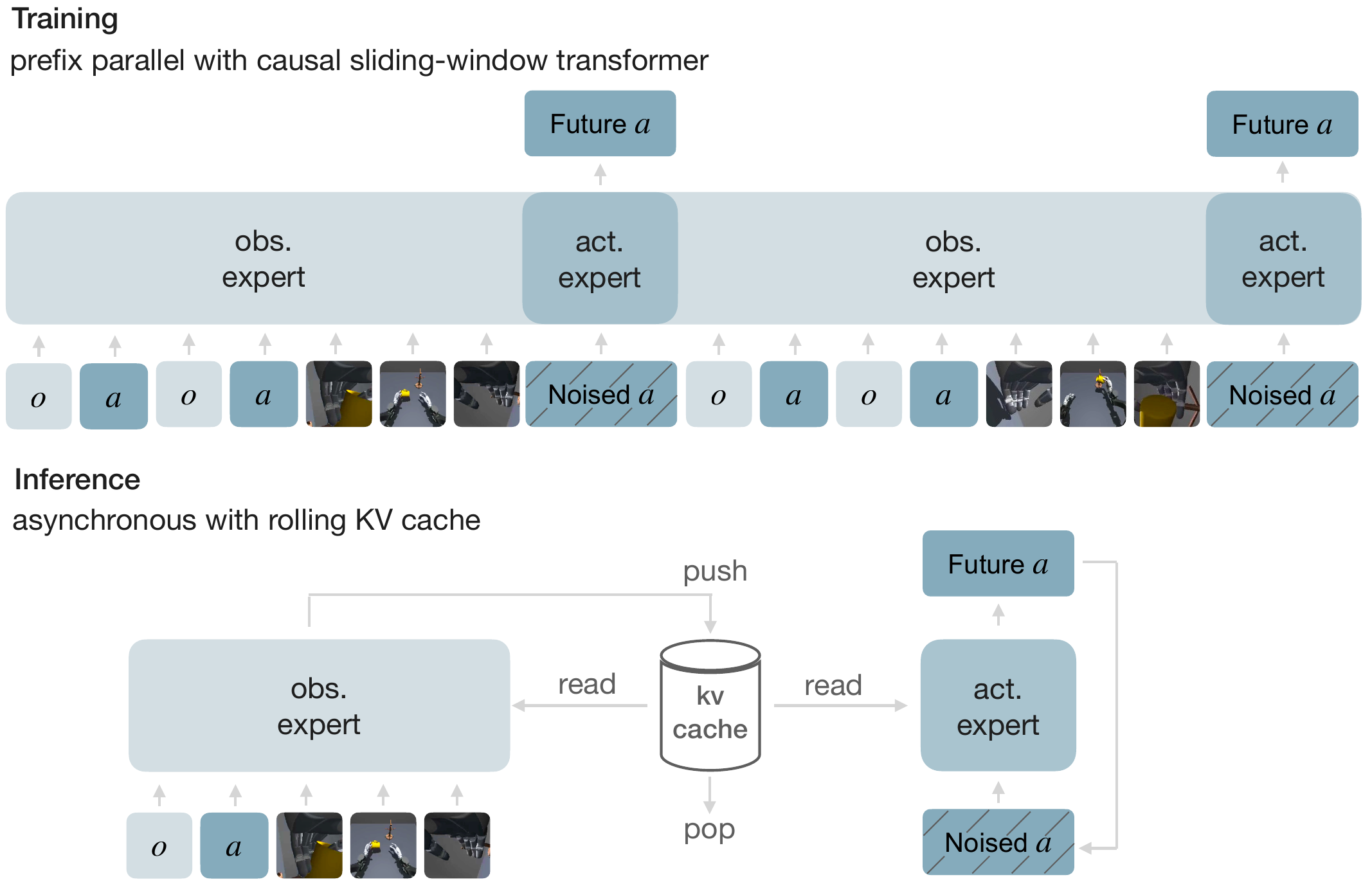}
    \caption{\textbf{Policy Architecture.}
    Our policy is a diffusion transformer that consumes a sequence of proprioception ($o$), action ($a$), multi-view images, and noised action chunks as interleaved tokens. In training, the model denoises all action chunks simultaneously with a causal mask, amortizing its sequence length. We additionally use sliding window attention in training, with a matched rolling KV cache for efficient inference.}
    \label{fig:SPD_arch}
\end{figure}

Our pre-training data spans diverse behaviors and low-level dexterous
primitives, so the policy must capture multimodal action distributions.
Following prior work on robot pre-training, we adopt a diffusion transformer that denoises action
chunks conditioned on visual and proprioceptive observations~\citep{pi0, abc2026}. Our data lacks dense language annotations and
broad scene coverage, so we forgo language conditioning. Instead, we
condition on the visuomotor  history, similar to~\cite{rpt}.

The model consumes an interleaved sequence of proprioception, action, and
visual tokens together with noised future action chunks, and supervises the
denoised chunks with a flow-matching velocity-prediction
objective~\citep{flowmatching}. Each training sequence spans 256 timesteps
recorded at 30\,Hz; to keep training efficient over this long context, the
model denoises all chunks in parallel under a causal mask, amortizing the
cost of processing the full sequence.

Visual observations from each camera
are encoded into patches by a frozen, pre-trained vision transformer and
pooled into a compact set of tokens via cross-attention from learnable
queries. To make this pooling context-dependent, the pooled visual tokens
cross-attend to the original ViT patches every two transformer blocks, as
in~\cite{pmlr-v139-jaegle21a}, and we subsample image inputs every 8
timesteps to remove redundancy between adjacent frames. Non-visual
modalities use modality-specific linear input and output projections over a
shared transformer trunk, and, as in~\cite{pi0}, the action-denoising expert
maintains its own set of weights.

All attention is causal, with tokens assigned temporal positions from their
timestamps via rotary embeddings. Noised action chunks additionally receive
absolute positional embeddings encoding both the flow-matching timestep and
the position within the chunk. To support a fixed-length KV cache at
deployment, every layer uses sliding-window attention with a 32-timestep
window. We train all models with Muon at a fixed learning rate of $10^{-3}$
and maintain an exponential moving average of the weights for inference.
 
\begin{figure}[t]
    \centering
    \includegraphics[width=\linewidth]{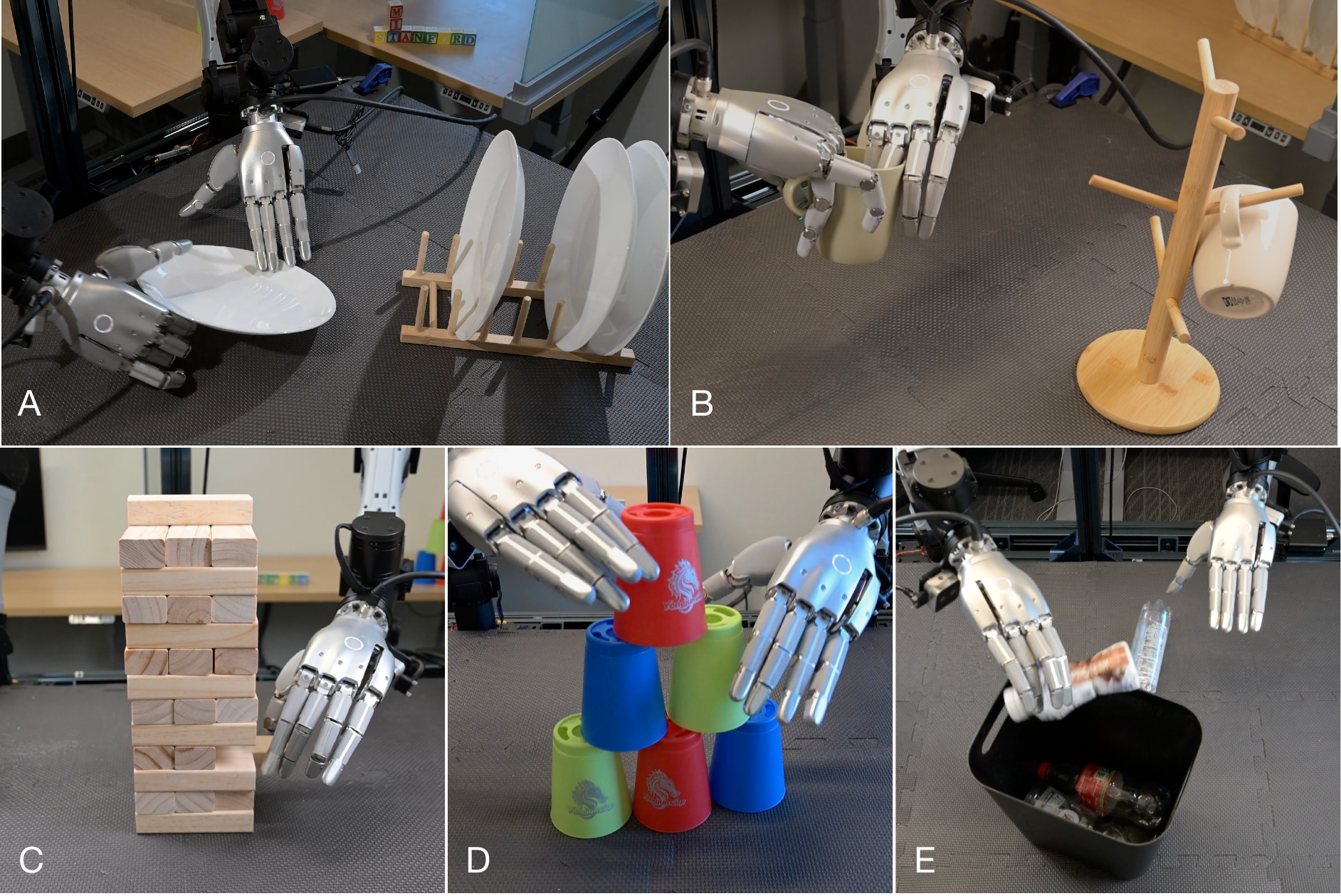}
    \caption{\textbf{Autonomous rollouts of five tasks.}
    (A) plate racking: lifting plates and racking them in a dish
    rack; (B) mug hanging: hanging a mug on a mug tree after a
    bimanual handover; (C) jenga playing: pushing a block out of
    the tower with one hand, pulling it free with the other, and placing
    it on top; (D) cup stacking: unstacking nested cups and
    building a pyramid; (E) bottles in bin: tossing bottles into
    a bin.}
    \label{fig:real-grid}
\end{figure}

\section{Experimental Results}
\label{sec:experiments}
We study the benefits of simulation pre-training by fully fine-tuning our policy on 1--2 hours of real-world demonstrations collected on our physical
robot per task. We evaluate on five tasks consisting of objects that are similar, but not identical to those seen in pre-training: (A) \texttt{plate racking}, (B) \texttt{mug hanging}, (C) \texttt{jenga playing}, (D) \texttt{cup stacking}, and (E) \texttt{bottles in bin}, shown in \Cref{fig:real-grid}. We include videos of each task on our \href{https://spd.bot/#videos}{project page}.

\begin{figure}[t]
    \centering
    \includegraphics[width=\linewidth]{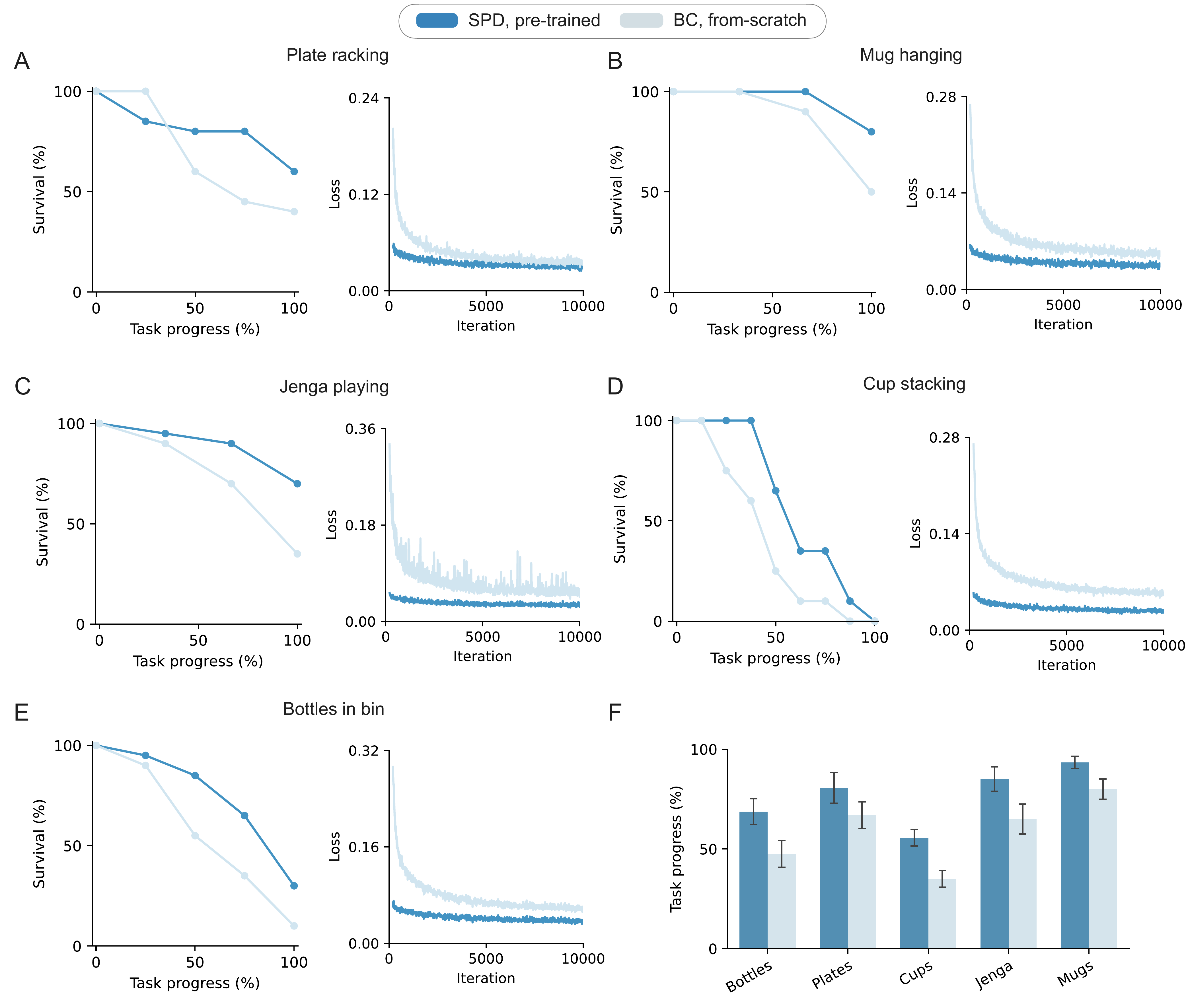}
    \caption{\textbf{Simulation pre-training improves real-world dexterity.}
    We evaluate five bimanual dexterous tasks on the real robot, comparing \abbr{} pre-trained checkpoints to those trained from scratch (BC). In (A-E), we report the fraction of trials that reach each successive manipulation stage and plot training loss curves. In (F), we plot average task progress; error bars show standard error.}
    \label{fig:results}
\end{figure}

\subsection{Benefits of Simulation Pre-training}

We compare our method against a baseline with the same architecture trained on real-world demonstrations of each task from scratch; results are in~\Cref{fig:results}. We perform 20 trials per checkpoint, and plot the fraction of trials that reach a certain amount of task progress.

On all five tasks, SPD reaches nearly every stage more often than the from-scratch BC baseline, and has higher average task progress. We also plot training loss curves for each task, which prior work shows to be positively correlated with downstream task performance for behavior cloning policies, unlike validation loss~\citep{abc2026}. The loss curves match our real-world observations, with \abbr{} checkpoints starting and converging to lower flow-matching loss values than from-scratch BC.

\subsection{Ablations on History Conditioning and Chunk Size}
\label{sec:arch_ablations}

\begin{figure}[t]
    \centering
    \includegraphics[width=0.8\linewidth]{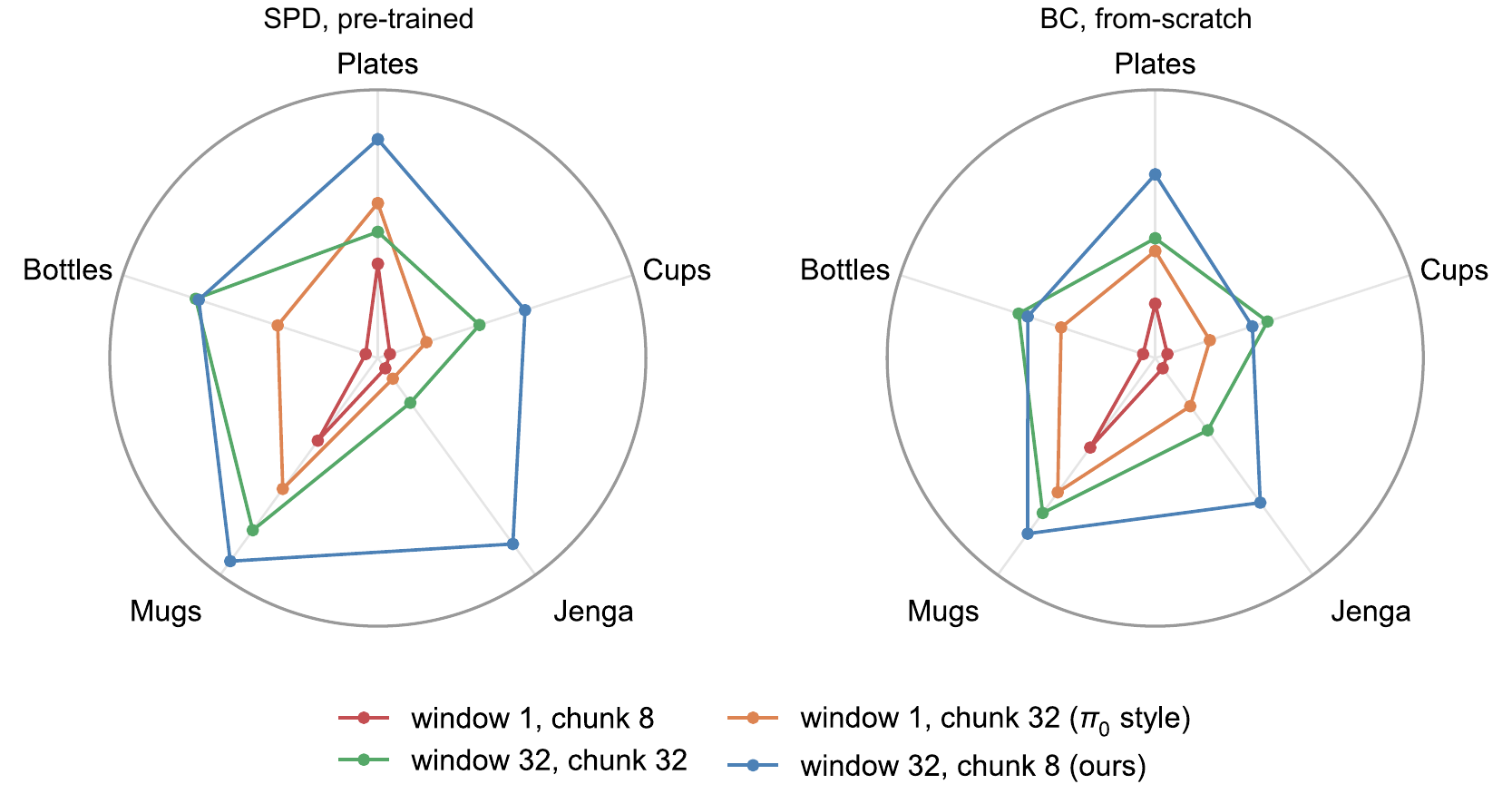}
    \caption{\textbf{Ablations on history conditioning and chunk size.}
    Visualization of the average task progress in \Cref{tab:ablation_full},
    for policies pre-trained with SPD (left) and
    trained from scratch (right).}
    \label{fig:ablation_radar}
\end{figure}

We study the roles of history conditioning and action chunking by sweeping
the sliding window size $w \in \{1, 32\}$ and the action chunk size $c \in \{8, 32\}$. Each of the four variants is trained
both from the pre-trained checkpoint and from scratch, sharing the
architecture, fine-tuning data, and training time of
\Cref{sec:policy_learning}, and is evaluated on all five tasks under the
protocol of \Cref{app:evaluation}. \Cref{fig:ablation_radar} plots
per-task progress for each variant, with the full numbers in \Cref{tab:ablation_full}.

Prior works, such as $\pi_0$~\citep{pi0}, train with a single frame of
context and a one-second action chunk, corresponding to $w = 1$, $c = 32$.
In this single-frame setting, we observe that reducing the
chunk size to $c = 8$ collapses performance, making the policy visibly shaky and less
temporally coherent. Adding history removes this trade-off: with a 32-step
window, the 8-step chunk becomes the strongest variant in both training
regimes, drawing temporal consistency from its context and reactivity from
its shorter chunk. This configuration also benefits the most from
pre-training: its average progress improves by 18 points over its
from-scratch counterpart, compared to 3 points or less for the other
variants.

\begin{table}[t]
    \centering
    \caption{\textbf{Average task progress for all variants.} Progress (\%) for every sliding-window ($w$) and action-chunk
    ($c$) combination, with the best variant per task bolded within each
    training regime. $w = 32$, $c = 8$ is the chosen configuration.}
    \label{tab:ablation_full}
    \begin{tabular}{lccccccc}
        \toprule
         & $w$ & $c$ & plates & mugs & jenga & cups & bottles \\
        \midrule
        \multirow{4}{*}{SPD, pre-trained}
         & \multirow{2}{*}{1} & 8 & 31.9 & 35.0 & 0.0 & 0.0 & 0.0 \\
         & & 32 & 55.6 & 58.3 & 5.0 & 15.0 & 36.2 \\
         \cmidrule(lr){2-8}
         & \multirow{2}{*}{32} & 8 & \textbf{80.6} & \textbf{93.3} & \textbf{85.0} & \textbf{55.6} & 68.8 \\
         & & 32 & 44.4 & 78.3 & 16.7 & 36.9 & \textbf{70.0} \\
        \midrule
        \multirow{4}{*}{BC, from-scratch}
         & \multirow{2}{*}{1} & 8 & 16.2 & 38.3 & 0.0 & 0.0 & 0.0 \\
         & & 32 & 36.9 & 60.0 & 18.3 & 17.5 & 33.8 \\
         \cmidrule(lr){2-8}
         & \multirow{2}{*}{32} & 8 & \textbf{66.9} & \textbf{80.0} & \textbf{65.0} & 35.0 & 47.5 \\
         & & 32 & 41.9 & 70.0 & 30.0 & \textbf{41.2} & \textbf{51.2} \\
        \bottomrule
    \end{tabular}
\end{table}

\section{Conclusion}

\label{sec:conclusion}

We present \algname~(\abbr), a simulation pre-training framework for dexterous manipulation. \abbr{} collects scalable, action-labeled, on-embodiment demonstrations in simulation without requiring physical robot hardware during pre-training. Pre-training a causal diffusion-transformer policy on this data improves real-world fine-tuning on five bimanual dexterous tasks compared to training from scratch. Our ablations show that history conditioning enables short, reactive action chunks and yields the largest gains from pre-training.

\textbf{Limitations.}
While \abbr{} improves real-world performance, it still depends on simulation scenes whose physics are tuned well enough for operators to produce realistic behaviors. If object masses, friction, or contact responses differ too much from real-world expectations, the collected demonstrations may encode strategies that transfer poorly. Our current pre-training data is also limited in scene and object diversity, and the real-world evaluation uses objects that are similar to those seen in simulation. Evaluating on more out-of-distribution objects, scenes, and task variations would better indicate how broadly the pre-trained policy generalizes.

\textbf{Future work.}
Simulation teleoperation is one of several routes to scaling dexterous pre-training, and a natural next step is to study how it complements the sources discussed in \Cref{sec:introduction} --- real-world teleoperation and egocentric human video --- within a mixed pre-training corpus. A second direction is scale: pre-training on more scenes, objects, and hours would allow characterizing the forms of generalization that emerge from simulation data itself. Finally, teleoperation is not the only way to generate experience in simulation. Reinforcement learning scales with compute rather than operator time, and our pre-trained policy offers a natural initialization for it.

\acknowledgments{We thank Scale AI for collecting the \texttt{spd-75h} dataset, Arthur Allshire and Ritvik Singh for sharing the ABC teleoperation infrastructure that was adapted for SPD, Sidney Nimako for his feedback, and Sharpa for their support. S.K. and A.R. are supported by the National Science Foundation (NSF) Graduate Research Fellowship Program. S.S. is supported by Meta and the NSF under Grant Numbers 2342246 and 2327974. This work was supported under project ID 43 as part of the Swiss AI Initiative, through a grant from the ETH Domain and computational resources provided by the Swiss National Supercomputing Centre (CSCS) under the Alps infrastructure. This work was also supported by a Packard Fellowship to P.I., ONR MURI grant N00014-22-1-2740, NSF Award 2153854, and Stanford HAI.}
 
\bibliography{references}

\newpage

\appendix
\section{Appendix}
\label{app:appendix}

\subsection{\texttt{spd-vr}: Simulation Teleoperation System}
\label{app:spd_vr}

\texttt{spd-vr} streams a MuJoCo~\citep{todorov2012mujoco} simulation to a Meta Quest~3 headset and maps
the operator's tracked hands onto the simulated robot. All computation runs on
a host workstation; the headset runs only a WebXR client that renders the
scene and reports hand poses. At load time the client receives every scene
body as a mesh, streamed in
binary at 60\,Hz, and sends back hand poses over a
USB tether. The simulation steps at 480\,Hz with the \texttt{implicitfast} integrator,
elliptic friction cones, and one no-slip iteration; control, streaming, and
recording run at 60\,Hz.

Collection is organized by a task registry: each task defines a
natural-language prompt, a target
duration, and a reset function that randomizes asset selection, object
placement, and physical properties; every sampled value is logged so episodes can be reconstructed
exactly. The operator controls recording with a three-button foot pedal
(checkpoint, pause, revert/skip); checkpoints are rejected while a hand is in
contact with an object, so reverting always restores a contact-free state where the operator can undo a mistake. After
collection, we cut spans with more than ten seconds of no hand--object
contact and render all clips in
parallel with a MuJoCo Warp adaptation of the Madrona batch
renderer~\citep{madrona} at $224\times168$, preserving instance segmentation
masks for the visual augmentations described in \Cref{sec:sim_data}.

\subsection{\texttt{spd-75h} Dataset}
\label{app:dataset}

\texttt{spd-75h} comprises 1{,}930 teleoperated simulation episodes
($\sim$75 hours) collected across six
scenes on the target bimanual dexterous embodiment. \Cref{tab:dataset_stats} lists the per-task episode and
duration breakdown.

\begin{table}[h]
    \centering 
    \caption{\textbf{\texttt{spd-75h} dataset statistics.} Per-task episode counts
    and durations, grouped by scene. Spelling variants are merged into a single
    \textsc{Spelling} task, and tasks with fewer than ten episodes are omitted.
    Durations are reported in minutes at 30\,Hz.}
    \label{tab:dataset_stats}
    \begin{tabular}{llrr}
        \toprule
        Scene & Task & Episodes & Minutes \\
        \midrule
        \multirow{6}{*}{Jenga} & Hollow tower & 92 & 567 \\
         & Tower & 87 & 473 \\
         & Dominos & 107 & 471 \\
         & Criss-cross & 103 & 386 \\
         & Handover (L$\to$R) & 96 & 172 \\
         & Handover (R$\to$L) & 72 & 109 \\
        \midrule
        \multirow{4}{*}{Spelling Blocks} & Spelling & 168 & 587 \\
         & Sort and unload & 25 & 144 \\
         & Pyramid & 32 & 136 \\
         & Sort vowels/consonants & 50 & 109 \\
        \midrule
        Mugs & Hang mug & 406 & 491 \\
        \midrule
        \multirow{2}{*}{Dishes} & Rack dishes & 129 & 285 \\
         & Plate dishes & 79 & 109 \\
        \midrule
        \multirow{3}{*}{Cups} & Pyramid & 44 & 109 \\
         & Stack two threes & 46 & 67 \\
         & Unstack & 30 & 47 \\
        \midrule
        Bottles & Toss in bin & 350 & 253 \\
        \midrule
        \textbf{Total} & & \textbf{1{,}916} & \textbf{4{,}516} \\
        \bottomrule
    \end{tabular}
\end{table}

\subsection{\texttt{spd-teleop}: Real-World Teleoperation System}
\label{app:spd_teleop}

\texttt{spd-teleop} mirrors the retargeting from \texttt{spd-vr}, and follows the system design from \texttt{abc}~\citep{abc2026} for hardware processes. The
stack is a set of single-purpose processes --- one per arm, hand, camera, and
logical component --- communicating over ZeroMQ publish--subscribe on local
IPC sockets, with a lightweight wire format of a JSON header plus raw array
bytes. Each process runs a fixed-rate loop: the leader streams commands at 60\,Hz, arm and hand followers servo
at 120\,Hz, and cameras publish at 30\,fps.

\emph{Wrists.} A WebXR client on the Quest headset streams controller poses
at 60\,Hz. A pedal press anchors the operator's wrist pose and the robot's end-effector
pose when tracking engages, after which motion is applied as relative deltas
with orientation changes re-expressed in the end-effector frame. Wrist
targets are solved using mink-based differential IK~\citep{mink2025}
(four QP iterations per tick with posture, joint-limit, and velocity tasks),
run in a dedicated process, with exponential smoothing and target
interpolation for jumps larger than 8\,cm.

\emph{Fingers.} Manus gloves provide 25 tracked keypoints per hand. For each
digit, the fingertip position in a palm-centric frame is mapped to a target
in the robot hand's palm frame by a per-operator affine map, fit by least
squares from a short calibration routine in which the operator holds
prescribed poses. The five fingertip targets
then drive mocap-based IK in a fixed-base simulation of the hand, whose resulting 22 joint angles are the hand command.

\emph{Robot control.} Arm commands are joint positions tracked by per-joint
PD gains with a gravity-compensation feedforward torque computed by MuJoCo
inverse dynamics on a model that includes the hand as end-effector payload;
motors are driven over CAN at 1\,Mbit/s with internal servo threads at
250\,Hz. Hands are position-controlled through the vendor SDK over Ethernet.
Episodes are recorded as per-stream HDF5 with parallel timestamp arrays:
camera frames as JPEG at native rate, arm observations at 120\,Hz, and commands at
60\,Hz, all resampled offline onto the 30\,Hz training grid.

\subsection{Real-World Evaluation}
\label{app:evaluation}

We evaluate each checkpoint with 20 trials per task from randomized initial
object placements. Each trial is scored against the per-task rubric in
\Cref{tab:rubrics}, and reported task progress is the achieved score
normalized by the task's maximum.

\begin{table}[h]
    \centering
    \caption{\textbf{Evaluation task rubrics.} Max score is the maximum
    achievable score for the task; reported progress is the achieved score
    normalized by this maximum.}
    \label{tab:rubrics}
    \small
    \begin{tabular}{p{2.4cm}p{2.8cm}cp{5.8cm}}
        \toprule
        Task & Setting & Max Score & Scoring Rubric \\
        \midrule
        \texttt{bottles in bin} & 4 bottles, 1 bin & 4 & \texttt{+1} for each bottle tossed into the bin. 60-second timeout. \\
        \texttt{plate racking} & 2 plates, 1 dish rack & 4 & \texttt{+1} for lifting each plate; \texttt{+1} for racking it \\
        \texttt{cup stacking} & 6 cups & 8 & \texttt{+1} for each cup placed correctly; \texttt{+1} for each subsequent destack move \\
        \texttt{jenga playing} & 1 tower & 3 & \texttt{+1} for pushing a middle block out; \texttt{+1} for pulling it from the other side without collapsing the tower; \texttt{+1} for placing it on top \\
        \texttt{mug hanging} & 1 mug, 1 mug tree & 3 & \texttt{+1} for lifting the mug; \texttt{+1} for the handover; \texttt{+1} for hanging it on the hook \\
        \bottomrule
    \end{tabular}
\end{table}

\subsection{Model Architecture}
\label{app:architecture}

Our policy is a 222M-parameter diffusion transformer that consumes an
interleaved multimodal token sequence and denoises future action chunks.
Each training sequence spans 256 timesteps at 30\,Hz. At every timestep the
sequence carries one proprioception token and one previous-action token
(both 56-D, normalized); every eighth timestep it additionally
carries four pooled tokens per camera and an 8-step noised action chunk.

\textbf{Vision pathway.} Images from the three cameras are encoded by a
frozen DINOv3 ViT-B/16~\citep{dinov3} into patch features. Per camera and
frame, four learnable queries pool the patch bank into four tokens via
cross-attention. To make the pooling context-dependent, the pooled tokens
re-attend to the raw patch bank through camera-specific cross-attention
blocks interleaved every two trunk blocks, as in~\citep{pmlr-v139-jaegle21a}.

\textbf{Trunk.} The trunk is an 8-block transformer with hidden size 768, 12
attention heads, and MLP expansion factor 4, with modality-specific linear
input and output projections. All attention is causal; tokens receive
temporal positions from their timestamps via rotary embeddings, and every
layer uses sliding-window attention over a 32-timestep window. Following $\pi_0$~\citep{pi0}, the action-denoising expert maintains its own 58M-parameter set
of weights, while observation tokens share the base trunk weights.

\textbf{Flow head.} Noised action chunks are constructed as
$x_t = (1-t)\,x_0 + t\,x_1$ with $x_0 \sim \mathcal{N}(0, I)$ and
$t \sim \mathcal{U}[0,1]$, and the model is supervised to predict the
velocity $v = x_1 - x_0$~\citep{flowmatching}. Each chunk token receives two
additive embeddings: a Gaussian Fourier embedding of the flow time $t$ and a
sinusoidal embedding of the position within the chunk, each mapped through a
two-layer MLP. During training, all chunks in the sequence are denoised in
parallel under the causal mask with independent per-chunk $t$, which
amortizes the cost of processing the 256-timestep context over 32 chunk
predictions. To reduce distribution shift from conditioning on history,
proprioception and action inputs are perturbed with i.i.d.\ Gaussian noise
($\sigma = 0.03$) during training.

\textbf{Inference.} At deployment the transformer runs as an incremental
engine over a rolling KV cache matched to the 32-timestep training window.
Each control tick appends the current observation tokens to the cache; on
chunk boundaries the engine integrates the flow ODE with 10 Euler steps and 
emits the next 8 actions.

\subsection{Training Hyperparameters}
\label{app:hyperparams}

\Cref{tab:hp_pretrain} lists the hyperparameters used for simulation
pre-training. \Cref{tab:hp_finetune} lists only the settings that differ during
real-world fine-tuning; all other hyperparameters are inherited from
pre-training.

\begin{table}[h]
    \centering
    \caption{\textbf{Pre-training hyperparameters.}}
    \label{tab:hp_pretrain}
    \begin{tabular}{ll}
        \toprule
        Hyperparameter & Value \\
        \midrule
        Batch size & 64 \\
        Learning rate & $1\times10^{-3}$ \\
        Learning rate schedule & constant \\
        Weight decay & 0.1 \\
        Optimizer & Muon (matrices), AdamW (rest) \\
        Parameters & 222M \\
        Parameters (vision encoder) & 86M \\
        Parameters (action expert) & 58M \\
        EMA half-life & 20 steps \\
        Training steps & 170k \\
        Sample rate & 30\,Hz \\
        Action chunk steps & 8 \\
        Image subsample steps & 8 \\
        Flow-matching noise schedule & uniform \\
        Observation noise & 0.03 \\
        Action noise & 0.03 \\
        Vision backbone & DINOv3 ViT-B/16 \\
        Vision queries & 4 \\
        \bottomrule
    \end{tabular}
\end{table}

\begin{table}[h]
    \centering
    \caption{\textbf{Real-world fine-tuning hyperparameters.} Only settings that
    differ from pre-training (\Cref{tab:hp_pretrain}) are listed; all other
    hyperparameters are inherited. Dataset size is reported per task.}
    \label{tab:hp_finetune}
    \begin{tabular}{lccccc}
        \toprule
         & {bottles} & {plates} & {cups} & {jenga} & {mugs} \\
        \midrule
        Training steps & 6k & 6k & 10k & 6k & 6k \\
        Dataset size (minutes) & 72 & 70 & 121 & 48 & 44 \\
        Dataset size (episodes) & 270 & 161 & 217 & 193 & 238 \\
        \bottomrule
    \end{tabular}
\end{table}

\end{document}